\documentclass[10pt,twocolumn,letterpaper]{article}

\usepackage[pagenumbers]{cvpr} 

\usepackage{graphicx}
\usepackage{amsmath}
\usepackage{amssymb}
\usepackage{booktabs}
\usepackage{multirow}

\usepackage[pagebackref,breaklinks,colorlinks]{hyperref}

\usepackage[capitalize]{cleveref}
\crefname{section}{Sec.}{Secs.}
\Crefname{section}{Section}{Sections}
\Crefname{table}{Table}{Tables}
\crefname{table}{Tab.}{Tabs.}

\def\cvprPaperID{6534} 
\def\confName{CVPR}
\def\confYear{2022}

\begin{document}

\title{Octree Transformer: Autoregressive 3D Shape Generation\\ on Hierarchically Structured Sequences}

\author{
{\parbox{\textwidth}{\centering \hspace{\stretch{2}} Moritz Ibing\hspace{\stretch{1.5}}Gregor Kobsik\hspace{\stretch{1.5}}Leif Kobbelt\hspace{\stretch{2}} }} \\
  Visual Computing Institute, RWTH Aachen University
}
\maketitle

\begin{abstract}
    Autoregressive models have proven to be very powerful in NLP text generation tasks and lately have gained popularity for image generation as well. However, they have seen limited use for the synthesis of 3D shapes so far. This is mainly due to the lack of a straightforward way to linearize 3D data as well as to scaling problems with the length of the resulting sequences when describing complex shapes. In this work we address both of these problems. We use octrees as a compact hierarchical shape representation that can be sequentialized by traversal ordering. Moreover, we introduce an adaptive compression scheme, that significantly reduces sequence lengths and thus enables their effective generation with a transformer, while still allowing fully autoregressive sampling and parallel training. We demonstrate the performance of our model by comparing against the state-of-the-art in shape generation.
\end{abstract}

\begin{figure*}[th]
    \centering
    \includegraphics[width=1.0\linewidth]{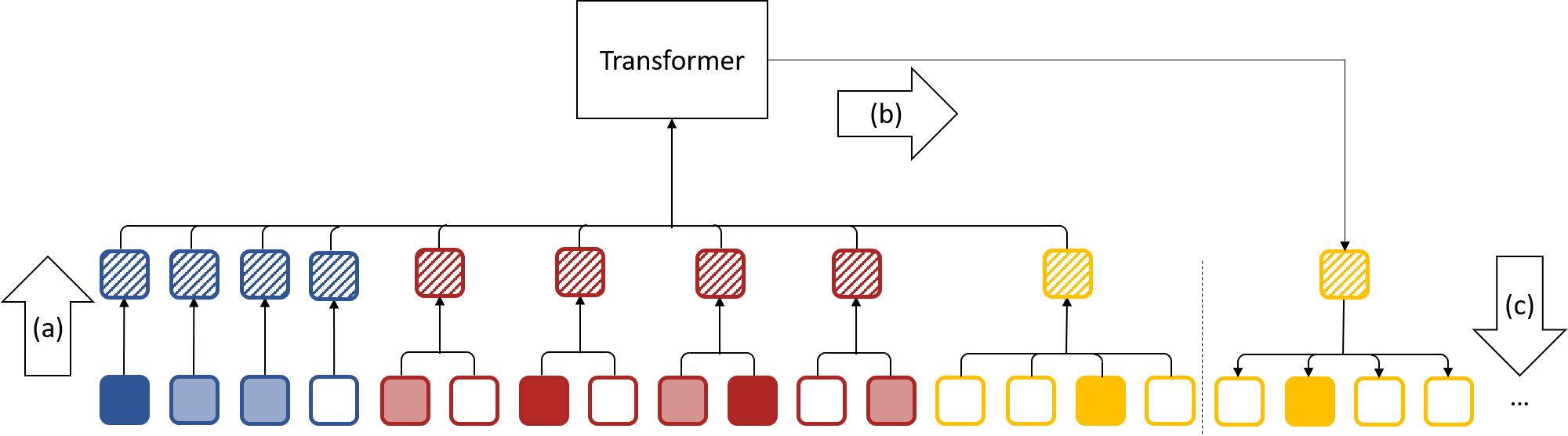}
    \caption{The three stages of our network: (a) sequence compression (b) autoregressive generation with transformer (c) sequence decoding}
    \label{fig:overview}
\end{figure*}

\section{Introduction}
Autoregressive models have become the standard paradigm for generating texts and have gained popularity for image generation as well. This is because they can explicitly model the likelihood of the data and thus can be trained to maximize it in a stable manner, a big benefit compared to \eg GANs.\\
A problem with applying this family of models to other fields, is that they can only generate sequences, as each generated element depends on all previous ones. This leads to issues when sequences become very long, as is the case \eg with images. When using recurrent networks, the training and sampling processes take a lot of time, as it can be done only sequentially. At least for the training this can be alleviated by the use of parallelizable models like CNNs \cite{van2016pixel,van2016conditional}, but here the receptive field is limited, meaning the generation of a new element can only depend on part of the previous sequence.\\
The advent of transformers \cite{vaswani2017attention,parmar2018image} somewhat alleviated this problem, as they can be trained in parallel and deal with much larger sequence lengths than CNNs. However, they cannot process arbitrary sequences, as the memory consumption grows quadratically with the length. There is ongoing research to reduce this memory requirement by approximating the attention matrix \cite{tay2020efficient}, but these approaches are agnostic to the underlying structure of the sequence.\\
When generating 3D data these problems are even more pronounced, as the trivial approach of sequencing a voxel grid (similar to images) would lead to a sequence length that grows cubically with resolution and quickly become intractable already at low resolutions. Thus, transformers have so far been seldomly adapted for the generation of 3D shapes and when they have, to different representations, like point clouds \cite{sun2020pointgrow} or meshes \cite{nash2020polygen}. These representations are difficult to work with, as they are either unstructured (point clouds) or irregular (meshes).\\
In our approach we do generate voxel grids, but to handle the cubic memory complexity, we use an octree representation, where the sequence length grows roughly quadratically with the resolution. Even then, a sequenced octree would still be too large to be processed with a transformer, requiring us to compress it further. \\
Compressing a sequence for transformer-based generation is not trivial, as in order to work on shorter sequences we not only need to define a compression method but we have to be able to expand this compacted sequence again in a fully autoregressive manner. This means we need to make sure, that not only the compressed embeddings depend on each other, but a token generated as part of a block in an expansion has to depend both on elements inside and outside of its block.\\
For this we make use of the octree structure, summarizing different subtrees with the help of convolutions. When generating new nodes of the tree, we again use convolutions for decompression, taking care, that the process stays fully autoregressive, while still being parallelizable and memory efficient.\\
The use of an autoregessive setup easily allows us to condition the generation process on class labels, whereas the octree encoding enables not only a structured compression scheme, but also the straightforward implementation of voxel superresolution. The earlier parts of an octree sequence represent a coarse shape representation while finer details are encoded in the later parts, thus superresolution can be formulated as sequence continuation.

\section{Related Work}

\paragraph{Transformer}
Autoregressive models in general and transformers in particular have often been used in the field of natural language processing \eg for text generation \cite{radford2019language, brown2020language}, as they are well suited for data that comes in the form of token sequences. More recently they have been applied to the field of vision as well to generate images. Early models here include PixelRNN, PixelCNN and the image transformer \cite{van2016pixel, van2016conditional, parmar2018image}. 
When using transformers to encode images Dosovitskiy \etal \cite{dosov2021animage} summarize blocks to reduce the sequence length and Wu \etal \cite{wu2021cvt} introduce convolutions for this compression. However, these are only employed on the encoding side and not in a generative context. Closest to our approach come and van den Oord \etal \cite{van2017neural} and Esser \etal \cite{esser2021taming} who compress an image by convolutions into a discrete set of codewords, which then can be generated in autoregressive fashion. However, here the compression can be seen as a preprocessing step and is not part of the generative model itself.

\paragraph{Octree}
Octrees \cite{meagher1982geometric} are a hierarchical spatial data structure that reduces the cubic memory complexity of voxel grids for storing 3D shapes, by recursively partitioning the space close to the surface, while saving information at a coarse resolution where it suffices. A downside is the higher complexity when implementing operations such as convolutions \cite{wang2017cnn,riegler2017octnet}, making their usage in neural network based processing relatively seldom. For generating 3d shapes with the help of octrees we are only aware of few methods, who all define convolutions on octrees and are either deterministic or based on VAEs \cite{hane2017hierarchical, tatarchenko2017octree, wang2018adaptive}. This formulation does not allow autoregressive generation as in our case.

\paragraph{3D generation}
Different neural networks have been developed to generate all kinds of 3D shape representations: point clouds \cite{nash2017shape,li2019point,achlioptas2018learning}, voxels \cite{sharma2016vconv, brock2016generative, wu2016learning}, functions \cite{chen2019learning} as well as those functions arranged in grid structures \cite{lim2019convolutional, ibing20213Dshape}. However, all of these employ latent variable models like VAEs or GANs to generate novel shapes. The realm of autoregressive models on the other hand is rather small. PointGrow \cite{sun2020pointgrow} and PolyGen \cite{nash2020polygen} generate point clouds and meshes respectively, using architectures based mainly on self-attention. However, as they do not use any sequence compression, they are limited in the size of point clouds or meshes, that they can represent.

\section{Octree Transformer}

Our network consists of three stages (\cref{fig:overview}): The encoding stage takes as input a sequenced octree (\cref{sec:octree}) and compresses it into a shorter sequence of latent vectors with an increasing compression factor for finer octree levels (\cref{sec:seq_encoding}). This sequence is used, to train a standard generative transformer decoder. We will not describe this transformer and rather refer to to Vaswani \etal \cite{vaswani2017attention}. The generated sequence of latent vectors produced by the transformer is then uncompressed and decoded into a sequence of octree nodes (\cref{sec:seq_decoding}).

\subsection{Octree Sequencing}
\label{sec:octree}
\begin{figure*}[tbp]
    \hfill
    \centering
    \begin{subfigure}[b]{0.4\textwidth}
        \includegraphics[width=1.0\linewidth]{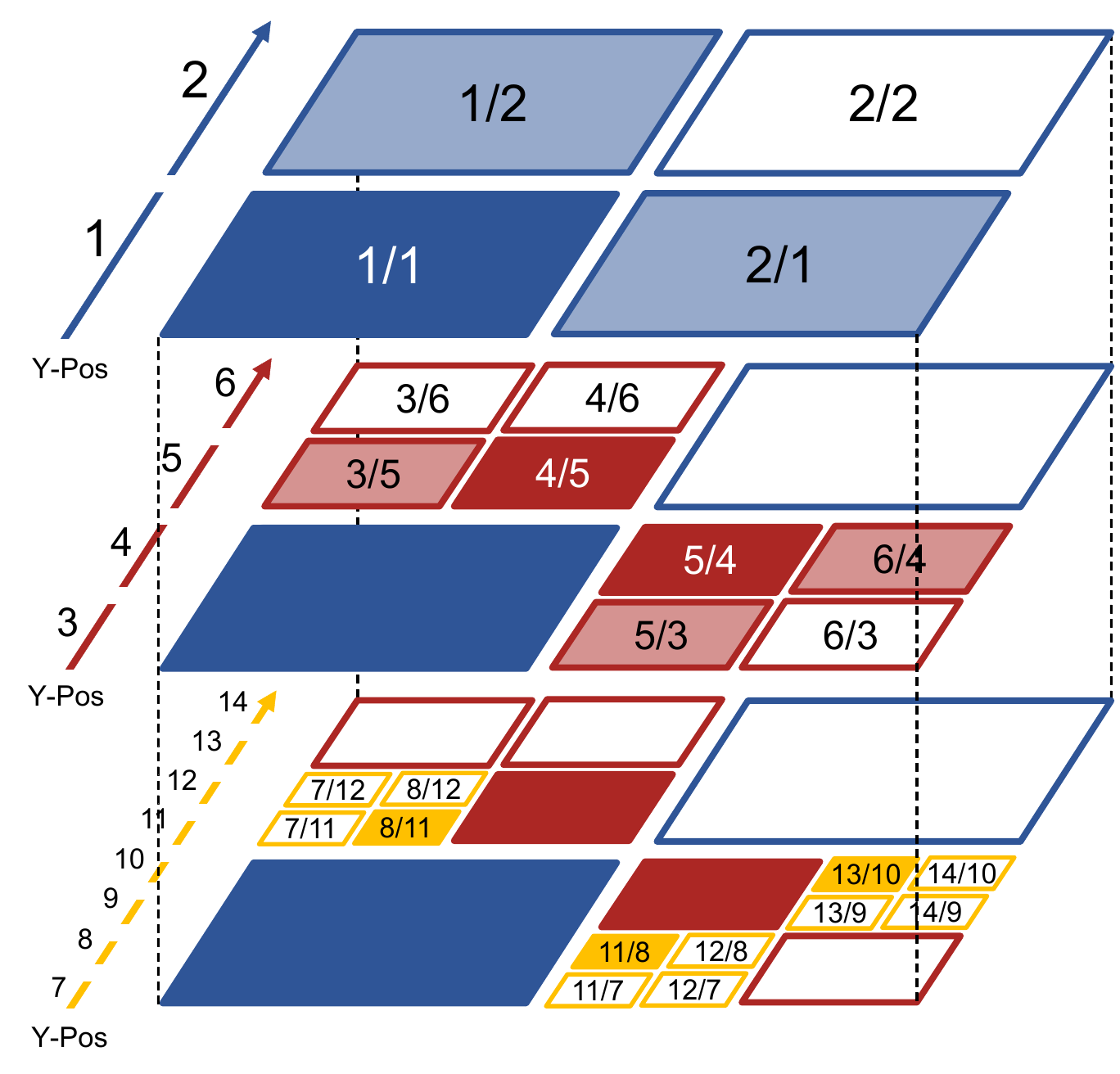}
        \caption{Shape}
        \label{subfig:octree_shape}
    \end{subfigure}
    \hfill
    \centering
    \begin{minipage}[b]{0.55\textwidth}
    \begin{subfigure}[b]{\textwidth}
        \includegraphics[width=1.0\linewidth]{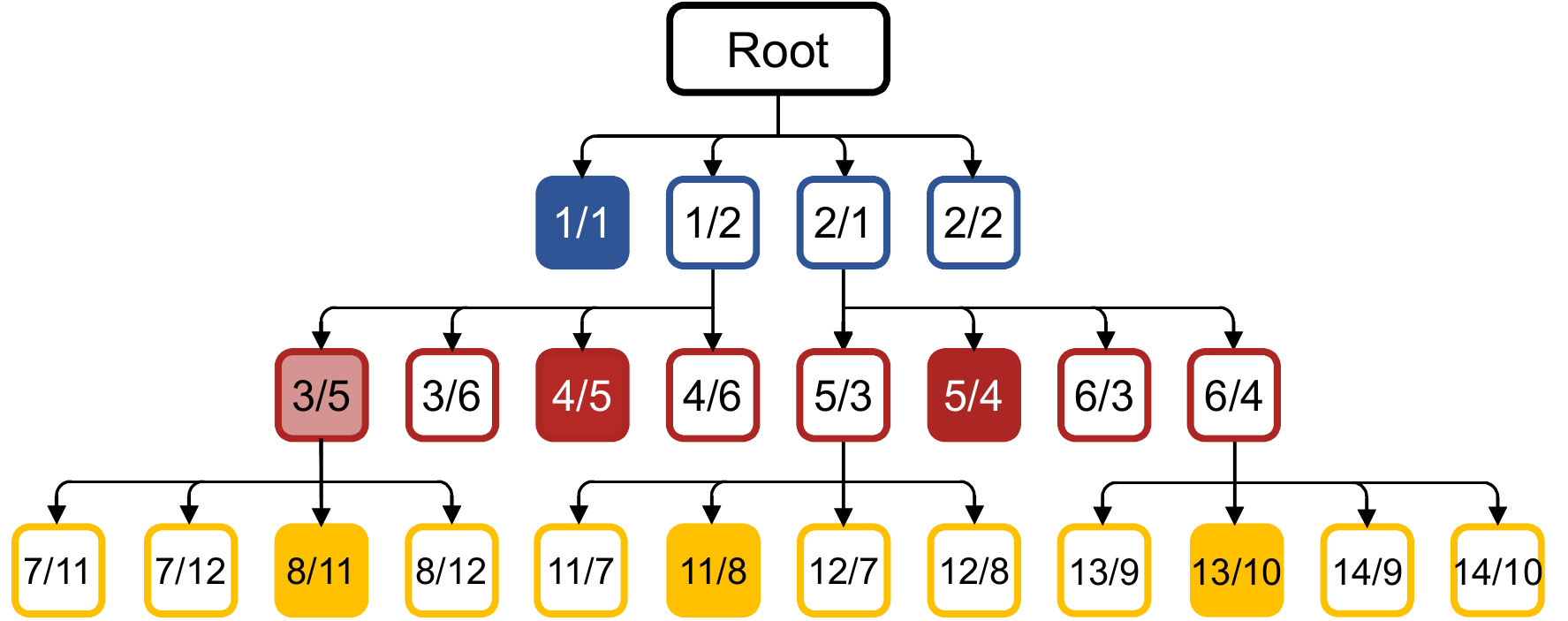}
        \caption{Quadtree}
        \label{subfig:octree_tree}
    \end{subfigure}\\ \\ \\
    \hfill
    \centering
    \begin{subfigure}[b]{\textwidth}
        \includegraphics[width=1.0\linewidth]{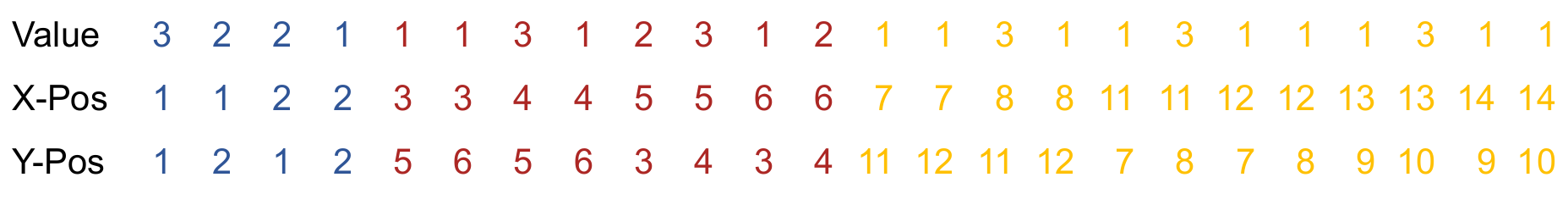}
        \caption{Sequence}
    \end{subfigure}
    \end{minipage}
    \caption{An example of our octree encoding (for simplicity we depict the encoding for a 2D quadtree). Position values are represented as X/Y for the x- and y-coordinates. Colors are used to indicate the depth level. Filled cells represent full (value=3), white cells empty (value=1) and transparent (value=2) cells mixed voxels. Below are the resulting value and positional sequences.}
    \label{fig:octree_encoding}
\end{figure*}

To construct an octree from a binary voxel grid, we start by setting up the bounding cube as the root cell. We then recursively subdivide a cell whenever it contains both empty and full voxels, splitting it into 8 cubes, called the cells "children". This process is stopped when all cell's contain only empty or full voxels (this happens at the latest, when we have reached the resolution of the original voxel grid). Each cell in the tree can thus be either empty, full or mixed (only intermediate nodes) (\cref{fig:octree_encoding}).\\
Next, this octree needs to be linearized into a sequence of tokens, from which the octree can be reconstructed. In the following we will discuss how we approach this problem. All examples in this section and the following will be on quadtrees, the 2D variant of octrees, for easier visualization. \Cref{subfig:octree_tree} shows the tree structure together with all the information we need from our octree.\\
Simply enumerating the values of each cell (empty (1), mixed (2), full (3)) in breadth-first manner would give us a sequence, that fully characterizes the octree and allows its reconstruction (since we know, that only mixed cells have children and these have exactly 8, we can infer exactly how many cells we have per layer). However, this way the spatial location for each cell can only be inferred from global context. Instead, we add a spatial encoding, that is unique over all depth levels. For this we enumerate all cells along each dimension individually from coarse to fine (\cref{subfig:octree_shape}), resulting in an unique id for each dimension. This spatial encoding replaces the 1D positional encoding usually employed, when working with transformers.\\
Together with the value of the cell $c$ we thus have 4 IDs (3 in 2D). For each of these we learn a discrete embedding $v(c), p_x(c), p_y(c), p_z(c)$. Note that the positional encodings do not need to be predicted by the generative model as they can be inferred from the values themselves.\\
As our positional encoding depends on the spatial location of a cell and not the position of its token in the sequence, it is not possible to infer the encoding of the next token without global context. We therefore embed each token as the sum of its own value and positional encoding plus its successor's positional encoding: $e(c_i) = v(c_i) + p_x(c_i) + p_y(c_i) + p_z(c_i) + p_x(c_{i+1}) + p_y(c_{i+1}) + p_z(c_{i+1}) $. This way when sampling a new token, the information of its position is easy to retrieve from its predecessor. The necessary context for this is always available, as the octree is generated from the root downwards.

\subsection{Sequence Compression}
\label{sec:seq_encoding}
\begin{figure}[tbp]
    \centering
    \includegraphics[width=1.0\linewidth]{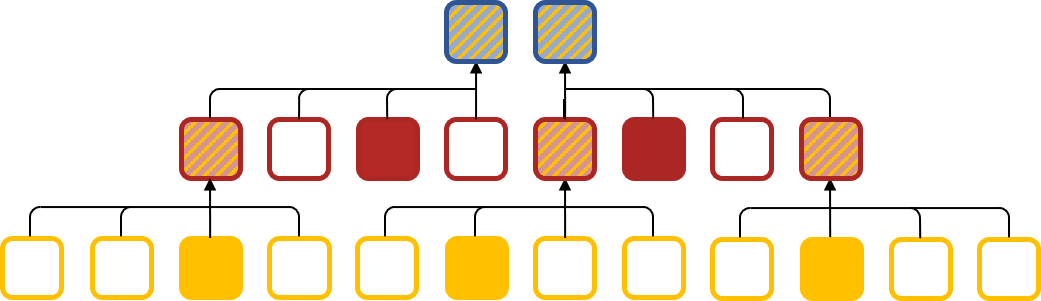}
    \caption{By compressing siblings we can achieve compression rates of up to 8 (4 in the quadtree example). For higher compression rates, we consider cousins (of higher order) by compression on the parent level. When compressing parent nodes we replace mixed tokens, by the representation of their children. For tokens that do not have any children (full/empty) we use their original embedding. The result of the compression is a latent vector for each compressed subtree.}
    \label{fig:octree_compression}
\end{figure}

Although we make use of an octree encoding to circumvent the cubic complexity of voxel grids, we still encounter prohibitively large sequence length when representing shapes in a high resolution. Therefore, we need to further compress our sequence in order to be able to fit the transformer's attention computation into memory. This compression should not treat our sequence simply as "flat" 1D data, but instead take the hierarchical octree structure into account. On the one hand this means, that not all tokens of the sequence should be treated equally, as cells from early depth layers are responsible for bigger parts of the shape (and are few), whereas later cells only encode details (and are many).
On the other hand the compression itself should take the structure of the octree into account and only compress cells together, that are spatially close, but not necessarily close in the sequence.\\
For compression rates of 2,4,8, this means that we combine the features of siblings in a tree. This can be done easily with strided convolutions (stride and kernel size equal to the compression factor).\\
For higher compression rates we encode entire subtrees into a single latent vector. For this we first compress all siblings and place their combined feature vector at the parent node. If this parent does have any siblings without children of their own, these will keep their own feature vector. We then repeat the compression at the parent generation (\cref{fig:octree_compression}).

\subsection{Sequence Decoding}
\label{sec:seq_decoding}
\begin{figure}[tbp]
    \centering
    \includegraphics[width=1.0\linewidth]{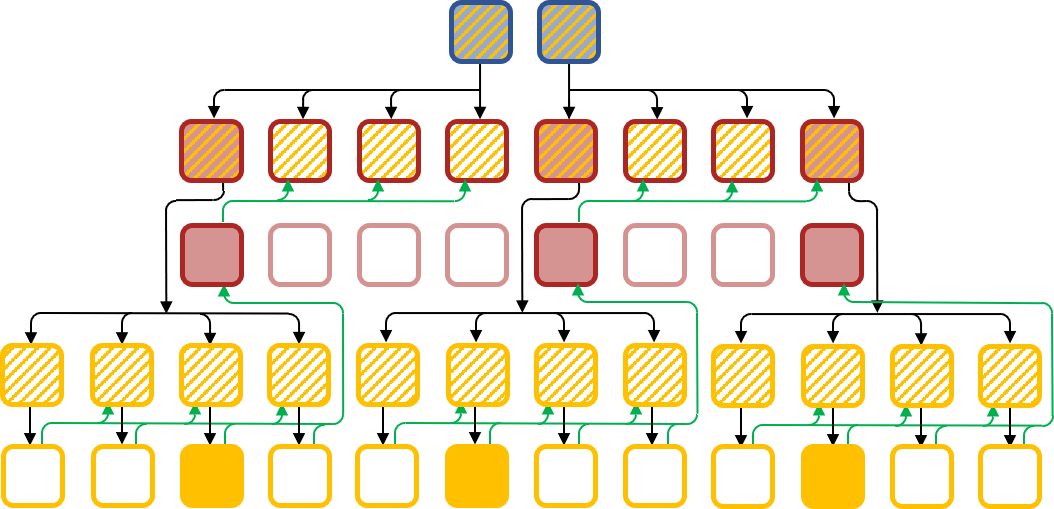}
    \caption{When generating a subtree from a sequence of compressed embeddings, we need to both upsample information from the embedding vectors (black arrows), as well as pass information from all previously generated tokens in the same subtree (green arrows).}
    \label{fig:octree_expansion}
\end{figure}

For generation we train the transformer decoder on compressed sequences, so that all feature vectors only depend on vectors positioned earlier in the sequence (when generating novel shapes, this sequence will be constructed one by one). However, to obtain a valid octree, we still have to undo the compression, and obtain logits from which to sample our cell information.\\
When no compression has been applied a linear layer with softmax non-linearity is enough to obtain logits.
When compression has been applied we have to undo it, using the reverse operations of what has been presented before.
Just doing a simple upsampling is not sufficient, as we would loose the autoregressive property of our generation. The probabilities of tokens generated this way would depend only on all previous tokens \textit{outside of the compressed subtree}. These dependencies are taken care of by the transformer. However, within the block the probabilities (of siblings and cousins) would not depend on each other. To model these dependencies efficiently we again make use of the octree structure (\cref{fig:octree_expansion}).\\
More concretely, we need to combine information from two different sources. Information about everything outside of the current subtree is encoded in the latent vector at the root and needs to be passed down to the leaves (\cref{fig:octree_expansion} black arrows), whereas information about already generated leaves within the tree needs to be passed to successor leaves on the same level (\cref{fig:octree_expansion} green arrows).\\
Let us start with the simple case of a compression factor $c \le 8$. We first need to undo the compression, by upsampling the latent vector, which simply can be done with a transposed convolution of the same stride and kernel size as was used for the compression, resulting in blocks of size $c$. Next, we need to model the dependency within these blocks. Each entry should depend on all entries occurring earlier in the sequence (within the same block). This can be done with what we call a masked block convolution. This passes information from one node to all successors within the same block. For this we just need to accumulate the respective latent vectors, which can be efficiently modeled with convolutions.\\
This operation can be seen as a sort of attention, where the attention weights are not input dependent, but instead only depend on the position. This is sensible in our case, as the position of our information (\textit{where} is the filled/mixed/empty cell positioned) contains more information than the value (which is only one of three options).\\
When we have compression factors higher than 8, we have to apply this strategy recursively. On the one hand the latent vector at the root is upsampled several times, taking the position of mixed tokens into account (only at mixed tokens we upsample further, as only mixed tokens have children). On the other hand information about leaf nodes needs to be propagated further. For this we pass it up a level (to the parent nodes) and there repeat the block convolution. Again, information is only passed to nodes later in the sequence. Nodes in the parent generation that do not have any children, do not contribute. This information is then added to the corresponding latent vectors in the same level and passed down again by the already described transposed convolution.\\
Note that we generally could model both the sequence encoding and decoding with additional smaller transformer networks as well. However, this would lead to a much larger memory consumption, counteracting the goal of the compression, which is why we have opted for this more lightweight approach.\\
Note that our compression algorithm is fully differentiable and we therefore can train the encoding, transformer and decoder together end to end.

\section{Evaluation}
\begin{table}[tbp]
    \centering

    
 


\begin{tabular}{cccccccc}

\toprule
Res     & Octree tokens & 0/4   & 0/8   & 1/8   & 2/8   \\
\midrule
2       & 8             & 2     & 1     & -     & -     \\
4       & 64            & 16    & 8     & 1     & -     \\
8       & 224           & 56    & 28    & 8     & 1     \\
16      & 816           & 204   & 102   & 28    & 8     \\
32      & 3352          & 838   & 419   & 102   & 28    \\
64      & 13264         & 3316  & 1658  & 419   & 102   \\
128     & 48530         & 12132 & 6066  & 1658  & 419   \\
256     & 144096        & 36024 & 18012 & 6066  & 1658  \\

\bottomrule
\end{tabular}
    \caption{Sequence lengths achieved with different compression schemes. The Octree tokens are from the 90 percentile per depth layer. The compression is denoted as $a/b$, where $a$ indicates the depth of the subtrees we collapse and $b$ how many tokens at the root level are encoded together.}
    \label{tab:compression}
\end{table}
All evaluations are performed on the ShapeNet Core dataset (v1) \cite{chang2015shapenet}. We use the training split from Chen and Zhang \cite{chen2019learning} (80\% training and 20\% testing) and randomly pick another 5\% from the training set for validation. The voxelized models are obtained from Hane \etal \cite{hane2017hierarchical}.\\
All experiments are done on a single RTX2080Ti. Our transformer decoder consists of 8 layers, uses 8 heads and a latent dimension of 256. We trained all models using the Adam optimizer \cite{kingma2014adam} and a training rate of $3\cdot10^{-5}$. Furthermore the first $10\%$ of the training steps are used for warm-up, where we linearly increase the learning rate from 0. We trained with a batch size of 1 for either 500 epochs on a single class or 100 epochs for class conditional models on the entire dataset. To increase sampling speed we used the fast transformer framework\cite{katharopoulos2020transformers,vyas2020fast}.
We sampled with a temperature of 0.8. We do not use an end-of-sequence token to stop sampling, but instead either stop when we have no more mixed tokens on the finest level, or when reaching a maximum resolution/octree level.\\
As we compare against GANs, which do not allow to compute a log-likelihood for test shapes, we adopt their evaluation scheme. For this we sampled a set of shapes 5 times the size of the test set, used the Light Field Descriptor \cite{chen2003visual} to compare single shapes and coverage (COV) and minimum matching distance (MMD) \cite{achlioptas2018learning} as well as edge count difference (ECD) \cite{chen2017new, ibing20213Dshape} to compare the respective sets.

\paragraph{Data Augmentation}
In all experiments we use data augmentation inspired by PolyGen\cite{nash2020polygen}, randomly scaling the shape independently along the different axes, with a piecewise linear warp. This augmentation is applied randomly during training. Note that even small changes in the voxel grid, can lead to big changes in the octree sequence, making this augmentation very important as shown in \cref{tab:ablation}.

\paragraph{Compression}
In \cref{tab:compression} we show which compression we achieve with a collection of different compression factors. From these numbers we can infer how much we need to compress in order to fit the sequence into memory, and what compression factor might be reasonable for what layer. In \cref{tab:ablation} we show different compression schemes.
After generating the octree, we filter out all sequences, that would have a length of more than 3400 after compression, in order to fit the training onto a single GPU. 

\paragraph{Loss Function}
Autoregressive models are usually trained by directly minimizing the negative log-likelihood, and we do the same here. However, we reason that in an octree representation getting the earlier elements in the sequence correct is much more important, than for later elements, as they cover bigger regions. We therefore weight the loss function according to the depth layer an octree element resides in. Starting with a weight of one for the first layer and then decrease by a factor of $\alpha$ for each subsequent layer. In order to be able to compare fairly, we normalize the weights to an average of one over a shape. In \cref{tab:loss} we evaluate different factors. To compare the results, we evaluate COV, MMD and ECD. As can be seen, differences are only marginal. For simplicity we therefore decided against any weighting.

\begin{table}[h]
    \centering
    \begin{tabular}{cccc}

\toprule
    \textbf{factor $\alpha$} & $\uparrow$COV & $\downarrow$MMD & $\downarrow$ECD \\
\midrule
 
1       & 76.47 & 2958 & 1889   \\
0.5     & 75.00 & 2921 & 1886   \\
0.25    & 74.78 & 2933 & 1767   \\
0.125   & 76.03 & 3011 & 2057  \\

\bottomrule
\end{tabular}
    \caption{Evaluation of different loss functions}
    \label{tab:loss}
\end{table}

\paragraph{Quantitative Results}

\begin{table}[b]
    \centering
    \begin{tabular}{lc}

\toprule
    \textbf{model} & bits/token \\
\midrule
uniform random              & 1.5850   \\
\midrule
full model                  & 0.0727   \\
- augmentation              & 0.1227   \\
+ later compression $A$     & 0.0762   \\
+ stronger compression $B$  & 0.0864   \\

\bottomrule
\end{tabular}
    \caption{Evaluating different design decisions. To put the numbers into context we report a baseline of uniform random sampling.}
    \label{tab:ablation}
\end{table}

\begin{table*}[tbp]
    \centering
    \begin{tabular}{cccccccc}
\toprule
                          &                   & \textbf{Plane} & \textbf{Car}   & \textbf{Chair} & \textbf{Rifle} & \textbf{Table} & \textbf{Average} \\
\midrule
\multirow{6}{*}{$\uparrow$\textbf{COV \%}} & 3DGAN \cite{wu2016learning}    &       & 12.13 & 25.07 & 62.32 & 18.80 &   \\
                          & PC-GAN \cite{achlioptas2018learning}            & 73.55 & 61.40 & 70.06 & 61.47 & 77.50 & 68.80 \\
                          & IM-GAN \cite{chen2019learning}                  & 70.33 & 69.33 & 75.44 & 65.26 & 86.43 & 73.36 \\
                          & Grid IM-GAN \cite{ibing20213Dshape}             & 81.58 & 80.67 & 82.08 & 81.47 & 86.19 & 82.80 \\
                          & Octree Transformer                              & 73.05 & 60.26 & 76.47 & 60.21 & 80.55 & 70.11 \\
\cmidrule{2-8}
                          & Train                                           & 85.04 & 85.67 & 84.73 & 84.00 & 87.13 & 85.13 \\
\midrule
\multirow{6}{*}{$\downarrow$\textbf{MMD}}  & 3DGAN                          &       & 1993  & 4365  & 4476  & 5208  &       \\
                          & PC-GAN                                          & 3737  & 1360  & 3143  & 3891  & 2822  & 2991  \\
                          & IM-GAN                                          & 3689  & 1287  & 2893  & 3760  & 2527  & 2831  \\
                          & Grid IM-GAN                                     & 3226  & 1225  & 2768  & 3366  & 2396  & 2607  \\
                          & Octree Transformer                              & 3664  & 1363  & 2958  & 3582  & 2496  & 2813  \\
\cmidrule{2-8}
                          & Train                                           & 2225  & 984   & 2317  & 3085  & 2066  & 2135  \\
\midrule
\multirow{6}{*}{$\downarrow$\textbf{ECD}}      & 3DGAN                      &       & 28855 & 26279 & 6495  & 32116 \\
                          & PC-GAN                                          &       &       &       &       &       \\
                          & IM-GAN                                          & 6543  & 20606 & 2553  & 3288  & 1018  \\
                          & Grid IM-GAN                                     & 355   & 1062  & 144   & 94    & 188   \\
                          & Octree Transformer                              & 2573  & 8563  & 1889  & 1835  & 1098     \\
\cmidrule{2-8}
                          & Train                                           & 1     & 11    & 1     & 2     & 5     \\
\bottomrule                          
\end{tabular}
    \caption{Comparison of different generative methods synthesizing 3D shapes. 
    Results from previous methods are taken from \cite{chen2019learning} and \cite{ibing20213Dshape}.
    Each subcategory was evaluated on a shape resolution of $64^3$ against a voxelized test set of the same resolution.
    3DGAN was not trained on the plane subcategory and no ECD values are reported for PC-GAN, thus some entries remain blank. For ECD no averages are reported, as values for different datasets are not comparable.}
    \label{tab:related_work_comp_64}
\end{table*}
In this section we want to justify several of the design decisions we made. For this we evaluate different configurations of our model and report the negative log-likelihood after training in \cref{tab:ablation}. For runtime reasons this ablation is done on the chair dataset for a resolution of 64.\\ 
As can be seen easily, data augmentation contributes a big factor in improving our results as it counteracts overfitting. In fact using data augmentation, we did not observe any overfitting at all, suggesting that the model size could be further increased given enough memory.\\
Lastly, we want to compare several choices regarding our compression scheme. As mentioned, the sequence can be compressed with a different factor at each depth level. Trying out every possible combination of compression is not feasible, therefore we restricted to three intuitive possibilities. As coarser levels are more important for reconstruction they should generally be compressed less than higher levels. However, at what rate the compression should increase is unclear. We evaluate two different schemes. In our baseline we increase the compression more or less linearly using factors of $(0/1, 0/1, 0/2, 0/4, 0/8, 1/4)$. In another experiment, we evaluate a "steeper" compression $A$. leaving earlier layers unchanged, and focusing the compression on the finer scales $A=(0/1, 0/1, 0/1, 0/1, 0/8, 1/8)$. The idea was, that this way we might lose information in the details, but keep the coarse structure better. Both compression schemes lead to roughly similar sequence lengths. As can be seen the former scheme is better able to reduce the loss, suggesting that a linear increase in compression is desirable (\cref{tab:ablation}). \\
Furthermore, we want to check if the compression scheme has any adverse effect at all, or if maybe a shorter sequence is even beneficial for training. For this we tested a stronger compression scheme $B=(0/1, 0/1, 0/4, 0/8, 1/4, 1/8)$ leading to sequences of roughly half the size. As can be seen, this leads to worse results. We conclude that the compression should be chosen so it leaves the sequence as long as memory permits.

\paragraph{Comparison}
After having found the best configuration for our model, we want to compare its abilities against other generative approaches. For this there are mostly GANs available. As their evaluation is carried out on a resolution of 64, we do the same for our model. We use the evaluation criteria described earlier in this section. The numbers for previous approaches are taken from Chen and Zhang \cite{chen2019learning} and Ibing \etal \cite{ibing20213Dshape}. Although we do not fully reach the state of the art, our method is competitive with the presented GANs (\cref{tab:related_work_comp_64}).

\begin{figure*}[h]
    \begin{subfigure}[b]{1.0\textwidth}
        \centering
        \includegraphics[width=1.0\linewidth]{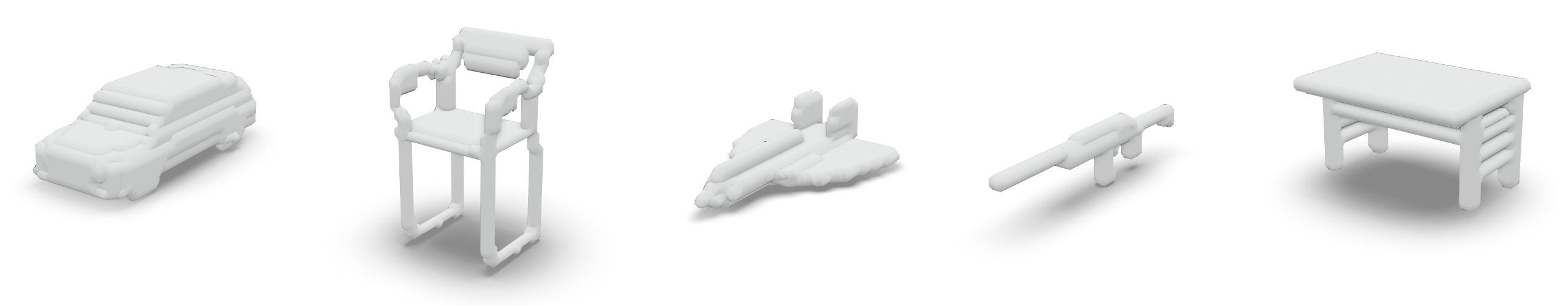}
        \caption{Resolution $64^3$.}
        \label{subfig:example_shapes_64}
    \end{subfigure}
    \hfill
    \begin{subfigure}[b]{1.0\textwidth}
        \centering
        \includegraphics[width=1.0\linewidth]{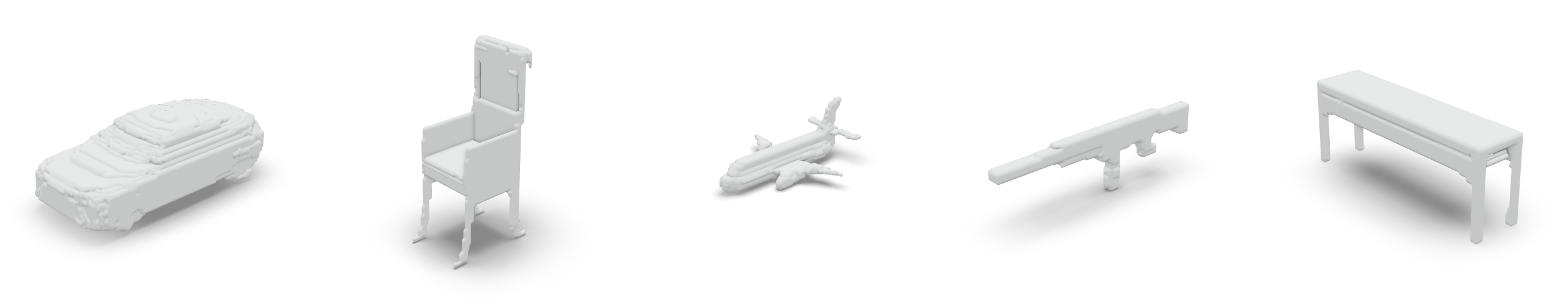}
        \caption{Resolution $128^3$.}
        \label{subfig:example_shapes_128}
    \end{subfigure}
    \hfill
    \begin{subfigure}[b]{1.0\textwidth}
        \centering
        \includegraphics[width=1.0\linewidth]{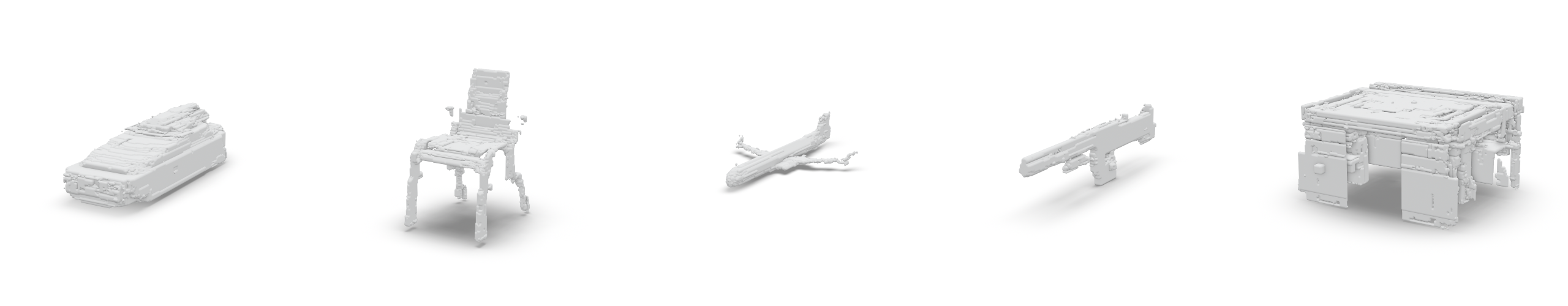}
        \caption{Resolution $256^3$.}
        \label{subfig:example_shapes_256}
    \end{subfigure}

    \caption{Samples synthesized by the class conditional Octree Transformer at different resolutions.}
    \label{fig:example_shapes}
\end{figure*}

We would have liked to compare against autoregressive approaches as well. But as the only available options PointGrow \cite{sun2020pointgrow} and PolyGen \cite{nash2020polygen} work on different shape representations have different input requirements and evaluated their model with different metrics on different subsets of the ShapNet dataset, we considered a fair comparison impossible.

\begin{figure*}[tbp]
    \begin{subfigure}[b]{1.0\textwidth}
        \centering
        \includegraphics[width=1.0\linewidth]{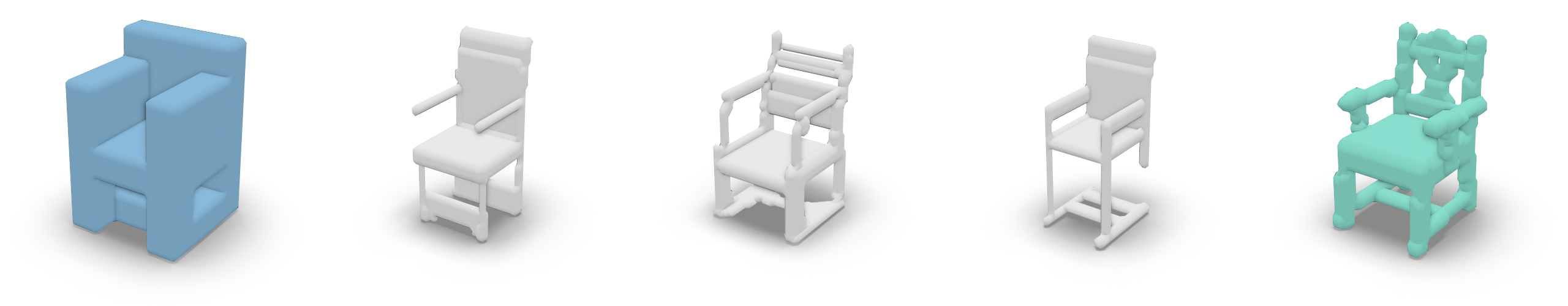}
        \caption{Resolution $64^3$.}
        \label{fig:example_shapes_64}
    \end{subfigure}
    \hfill
    \begin{subfigure}[b]{1.0\textwidth}
        \centering
        \includegraphics[width=1.0\linewidth]{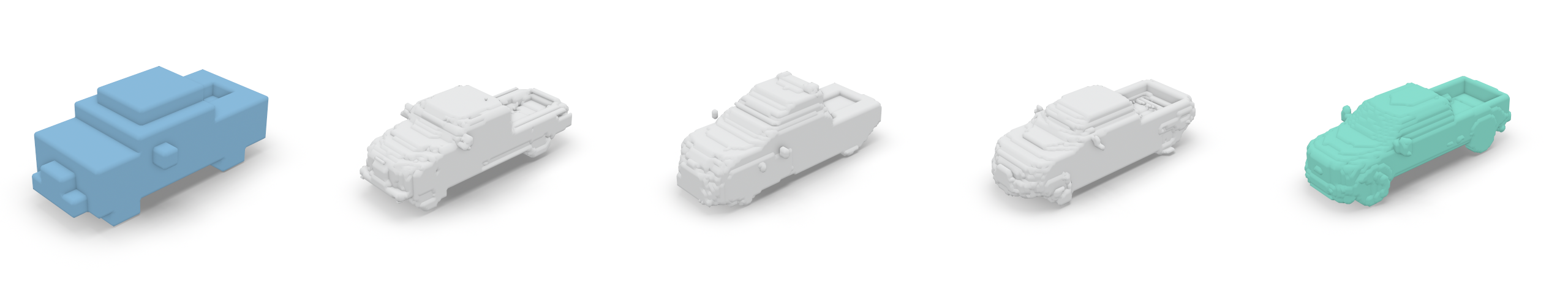}
        \caption{Resolution $128^3$.}
        \label{fig:example_shapes_128}
    \end{subfigure}
    \hfill
    \begin{subfigure}[b]{1.0\textwidth}
        \centering
        \includegraphics[width=1.0\linewidth]{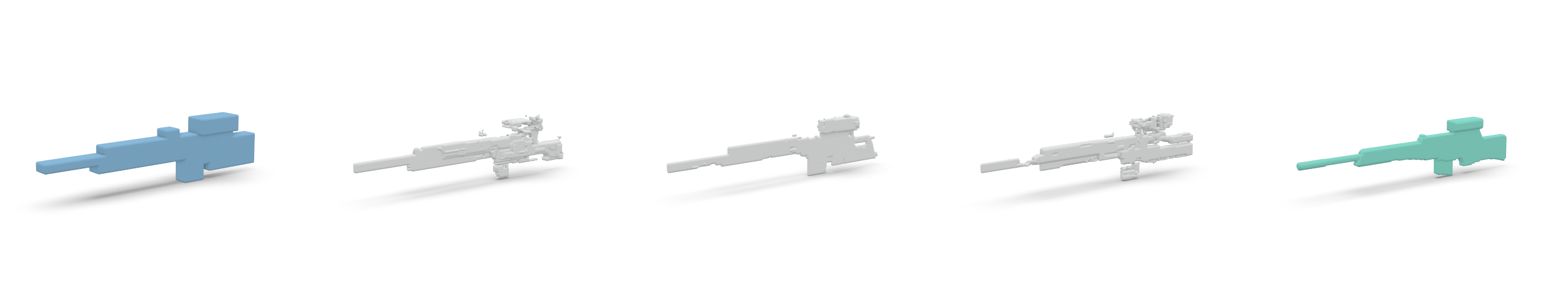}
        \caption{Resolution $256^3$.}
        \label{fig:example_shapes_256}
    \end{subfigure}

    \caption{Superresolution samples synthesized by the Octree Transformer at different resolutions. Our model computes an upsampling of factor eight. The leftmost shape is the preconditioning and the rightmost shape the actual model from the dataset. In grey we show three different upsamplings.}
    \label{fig:superres_shapes}
\end{figure*}

\paragraph{Qualitative Results}
Here we show samples generated by our approach. In \cref{fig:example_shapes} we have shapes generated at different resolutions sampled from class conditional models. For each resolution we trained a different model (although in principle higher resolution models can generate lower resolutions as well). To fit the models into memory we used different compression strategies. The model for resolution 64 is the same as presented earlier as our baseline. For a resolution of 128 we use a compression of $(0/1, 0/1, 0/4, 0/8, 1/4, 1/8, 2/4)$ and for resolution 256 $(0/1, 0/1, 0/4, 0/8, 1/4, 1/8, 2/4, 2/8)$. We achieve our best results at a resolution of 64 and can see a decrease in quality at higher resolutions. This might be due to the higher compression factor that is necessarily employed, as well as due to the limited size of our model (a bigger network with more heads might be able to focus better on the coarse shape as well as on fine details). Furthermore, our hyper-parameters were optimized for a resolution of 64.\\
Next, we show results for the superresolution task in \cref{fig:superres_shapes}. Here, we provide the model with truncated sequences from the test set, describing the shape only up to a given resolution and let our model complete these sequences. Again, we show results at three different resolutions using the same models as in the last task. In all examples we increased the resolution by a factor of 8 (3 levels). We show the input shape, some generated samples, and the original shape from the test set. At high resolutions, we see the same quality problems, together with a lack of diversity (compared to \eg resolution 64), that can be explained by the more extensive precondition and the comparatively small dataset.

\paragraph{Limitations}
The quadratic memory complexity assumed to be obtainable by using an octree structure only holds for natural shapes, where the surface is not too complex. In the worst ("fractal") case, where even in the finest resolution every voxel is near the shape boundary, the octree does not achieve any memory reduction at all (we even increase the memory by a factor of 4/3, due to the overhead of saving the tree). However, in practice we never encountered such a case. Still, the length of the octree sequence varies strongly with shape complexity.\\
Even though our compression scheme is fully parallelizable and decreases the effective sequence length for the transformer, during sampling we still need to sample each token of the uncompressed sequence one after the other. We do not need to apply the transformer for each token, but still this can lead to long sampling times (several minutes) for complex shapes at high resolution.\\
Lastly, the quality of our generated shapes is not always as high as desired. Especially when generating shapes at high resolutions, we sometimes produce noisy or incoherent shapes.

\section{Conclusion}
In this work we introduced a new deep generative model, based on the transformer architecture for the generation of 3D shapes in the form of octrees. In order to deal with long sequences we developed a compression scheme, that significantly reduces sequence length, while still allowing fully autoregressive generation. Our model can be applied for the generation of novel shapes, as well as for increasing the resolution of existing ones.\\
Since our method still produces voxel grids (in high resolution) it would be interesting to see if it is possible to generalize from binary grids to more smooth or expressive representations like implicit functions. A possible direction of research would be to save more information in the voxel cells. \\
Furthermore, our convolutional compression scheme might be of interest in other fields that deal with long sequence lengths, like the domain of image generation.

{\small
\bibliographystyle{ieee_fullname}
\bibliography{egbib}
}

\end{document}